# Scalable Dynamic Tactile Sensing Enabled by Passive and Flexible Acoustic Waveguides

**Guimin LONG, Changhong LINGHU, Chuanping LIU, Ke XU, Xingjian JING*,**

Department of Mechanical Engineering, City University of Hong Kong, Hong Kong, China

**Corresponding Author: Xingjian JING,** City University of Hong Kong, Hong Kong, China;

Email: xingjing@cityu.edu.hk

## Abstract

Artificial dynamic tactile sensing requires sensitivity, robustness, and compliance, yet existing technologies face trade-offs when scaling to large-area arrays, compounded by wiring complexity and cost. Here, we report a passive distributed paradigm using deep sub-wavelength acoustic waveguides that decouples performance from structural flexibility. Elastic-membrane-capped Helmholtz resonators interconnected by spring-reinforced microtubes form an enclosed network with invariant acoustic transmission under macroscopic bending. By sparsely embedding microphones, the system achieves real-time localization (4 mm highest spatial resolution; >99% accuracy in a 4 microphones 64-node sensing array) and waveform reconstruction of low-frequency signals (<100 Hz). Fast Continuous Wavelet Transform and a lightweight neural network enable inference within 5.5 ms. We demonstrate conformable prototypes—fingertip arrays, a tactile glove, and large-area skins—detecting stimuli from single-hair contact to 5-mg particle impacts, arterial pulse waves, feather touches, and finger contact. This establishes a scalable, flexible, low-cost paradigm for next-generation human-machine interfaces.

## Introduction

Dynamic tactile signals encode both frequency and amplitude information, making them particularly valuable in human–machine interaction applications [1-3]. Human tactile perception primarily relies on subcutaneous mechanoreceptors, which can be broadly classified into fast-adapting and slow-adapting corpuscles [4, 5]. The fast-adapting receptors—Meissner and Pacinian corpuscles—are chiefly responsible for detecting vibratory stimuli ranging from a few Hertz to several hundred Hertz [6, 7], highlighting the critical importance of this frequency band for artificial dynamic tactile sensing systems.

In recent years, significant research efforts have been devoted to developing artificial tactile perception systems [8-12]. A wide range of sensors based on resistive [13-16], capacitive [17, 18], inductive [19, 20], electromagnetic [21, 22], piezoelectric [23-25], triboelectric [26-28] and vision or optical [29-31] principles have been developed for human-machine interaction applications [9, 32]. However, when deployed in large-scale arrays, many of these systems face challenges including excessive wiring requirements [23, 33] and increased cost, often accompanied by compromises in spatial resolution, accuracy, robustness and sensitivity. These limitations not only heighten system complexity but also undermine durability, cost-effectiveness, and practicability.

To mitigate wiring complexity, some studies have adopted an M (rows) + N (columns) addressing strategy [23, 34-36], which achieves stimulus localization by decoupling signals across orthogonal channels. Although this approach offers simpler wiring compared with traditional dot-matrix architectures, the number of connections in large-scale arrays remains substantial. Furthermore, row-column addressing schemes typically require stacking multiple transducer layers, which leads to more complex fabrication processes and higher manufacturing costs.

A notable alternative is distributed sensing, typically implemented using optical [37-41] or ultrasonic [42-44] waveguides. In such systems, optical fibres or other waveguiding media are embedded in the sensing region, and external stimuli are inferred from perturbations induced in the guided optical or ultrasonic waves. Compared with the row-column strategy, distributed sensing substantially reduces the number of connection endpoints and offers exceptionally lower incremental cost when scaling to larger sensing areas.

Nevertheless, optical and ultrasonic waveguide-based sensing systems generally operate in an active interrogation manner. Their reliance on externally generated carrier waves (light or ultrasound) and complex demodulation hardware makes these systems bulky and expensive, limiting their applicability in portable or low-cost scenarios [45]. In addition, their responses are highly sensitive to device deformation and installation conditions [46], rendering them more suitable for detecting static or ultra-low-frequency (few-Hertz) physiological signals [47, 48], deformations [49] or strains [43, 50-53]. To the best of our knowledge, no distributed waveguide system has been specifically demonstrated for broadband dynamic tactile sensing. Furthermore, ultrasonic signals experience strong attenuation over long waveguide distances and are susceptible to reflections in curved geometries [51], which further limits their versatility in practical applications.

By contrast, low-frequency acoustic waves exhibit lower transmission loss and propagate more effectively through nonlinear or curved waveguides, resulting in robust and stable signal characteristics that are advantageous from an engineering standpoint. However, their use in tactile or wearable sensing has remained largely unexplored. Several factors likely account for this gap. First, conventional acoustic waveguide-based distributed sensing typically follows an active paradigm, in which external stimuli are inferred from perturbation imposed on an artificially generated carrier wave. Within this framework, the inherent stability of low-frequency acoustic waves makes them unsuitable as carrier signals and they are therefore usually excluded during system design. By contrast, a passive framework—where stimuli themselves generate the acoustic signals of interest—does not require a stable carrier, thereby opening a pathway for utilizing low-frequency acoustic responses directly. Second, according to the diffraction limit commonly associated with the Rayleigh criterion, the theoretical spatial resolution for acoustic source localization is approximately half the wavelength. The long wavelengths associated with low-frequency acoustics thus intrinsically limit achievable spatial resolution [54], at least within conventional beamforming or time-of-flight localization schemes. Third, under the time–frequency uncertainty principle [55], commonly used real-time time–frequency analysis methods --- such as the Short-Time Fourier Transform (STFT) --- require long temporal windows to analyze low-frequency components, resulting in poor time resolution and making real-time feature extraction challenging. This limitation is even more pronounced in source localization tasks, which typically depend on multiple frames of time-frequency information [56-58]. However, because these three constraints arise from the active-sensing paradigm, conventional spatial sampling, and classical spectral analysis, respectively, they are not fundamental physical barriers to all possible system architectures.

In this article, we introduce a passive distributed dynamic-stimulus sensing framework based on deep-subwavelength acoustic waveguides. The system comprises elastic-plate-capped Helmholtz resonators interconnected by spring-reinforced flexible microtubes with deep-subwavelength effective dimensions (**<0.1λ**), enabling resonator arrays to be configured in diverse layouts. Unlike conventional localization methods that rely on wavelength-limited spatial sampling, this tailored waveguide architecture produces distinct acoustic signatures when stimuli are applied to different resonators. Because each resonator acts as a unique spatial encoder, the effective spatial resolution is determined by the distinguishability of these spectral fingerprints rather than by the acoustic wavelength alone, thereby circumventing the conventional diffraction limit. By sparsely embedding only a small number of microphones within the enclosed waveguide network, the system achieves high-spatial-resolution localization (4 mm) of low-frequency (<100 Hz) stimulus-induced acoustic responses. For signal processing, we employ the fast Continuous Wavelet Transform (fCWT) [59] for efficient time–frequency analysis and integrate deep learning architectures [60] for spectrum-based source localization. Owing to the inherent robustness of deep-subwavelength acoustic waveguides to tube curvature, sensing performance is effectively decoupled from structural flexibility. Importantly, the original external stimulus can also be reconstructed using a pre-calibrated numerical waveguide transfer matrix.

To the best of our knowledge, this work presents the first passive distributed waveguide sensing framework capable of high-resolution dynamic tactile detection through deep subwavelength acoustic waveguides, challenging long-standing assumptions about the intrinsic limitations of low-frequency acoustics. By uniting a tailored waveguide architecture with sparse transducer deployment and real-time localization algorithms, we establish a scalable, flexible, and ultra-low-cost platform for next-generation human–machine interfaces. Operating in a previously unexplored regime, the system enables precise localization and reconstruction of low-frequency stimuli while maintaining high structural flexibility, offering a promising pathway toward scalable and affordable tactile technologies.

# Results

## Mechanism of deep-sub-wavelength waveguide dynamic tactile sensing

The typical architecture of conventional large-scale tactile sensing arrays is schematically illustrated in Fig. 1a. In contrast, this work propose a purely passive sensing paradigm, as shown in Fig. 1b. Here, acoustic resonators capped with elastic top plates are interconnected in series via spring-reinforced flexible microtubes, with two microphones positioned at the array termini. When an external stimulus contacts the top plate of a resonator, it generates an acoustic signal within the resonator chamber. This signal propagates through the series-connected waveguide network and is captured by the two embedded microphones.

Because the physical dimensions of the microtubes and resonators are far smaller than the acoustic wavelength, the system operates in the deep-subwavelength regime. In this regime, the transmitted acoustic waves propagate as quasi-plane waves (Fig. 1c) and are largely immune to tube bending [61], thereby preserving sensing fidelity in serpentine or coiled configurations. Since acoustic signals originating from different sensing points (resonators) traverse distinct waveguide paths before reaching the two microphones, they exhibit characteristic inter-channel differences. A lightweight neural-network classifier exploits these disparities to achieve accurate stimulus localization. Furthermore, by incorporating calibrated transfer functions, the system enables precise reconstruction of the original dynamic stimulus. Owing to the computational efficiency of both the localization and reconstruction algorithms, the platform supports real-time perception across a broad spectrum of dynamic tactile events.

Benefiting from the complete decoupling between waveguide characteristics and structural deformation, the one-dimensional dual-microphone sensing array in Fig. 1b can be reconfigured into a two-dimensional flexible tactile array (Fig. 1d) via one-dimensional coiling. Moreover, because the system monitors the acoustic field induced by external excitations within the enclosed waveguide space, it affords diverse topological configurations by varying the number of microphones, as illustrated in Fig.-1e. For a system with $M$ microphones, the minimum and maximum numbers of series-connected resonator lines are $M-1$ and $M(M-1)/2$, respectively, yielding a total of $[M(M-3)+4]/2$ possible configurations. Assuming each line contains *N* sensing points (resonators), the maximum number of sensing points in an *M*-microphone system is $N \times M(M-1)/2$. Consequently, the sensing-point density per transducer—a key metric of transducer efficiency—is $N(M-1)/2$ times that of conventional one-transducer-per-point arrays, and $(M-1)(M+N)/(2M)$ times that of the row–column strategy (Fig. 1f).

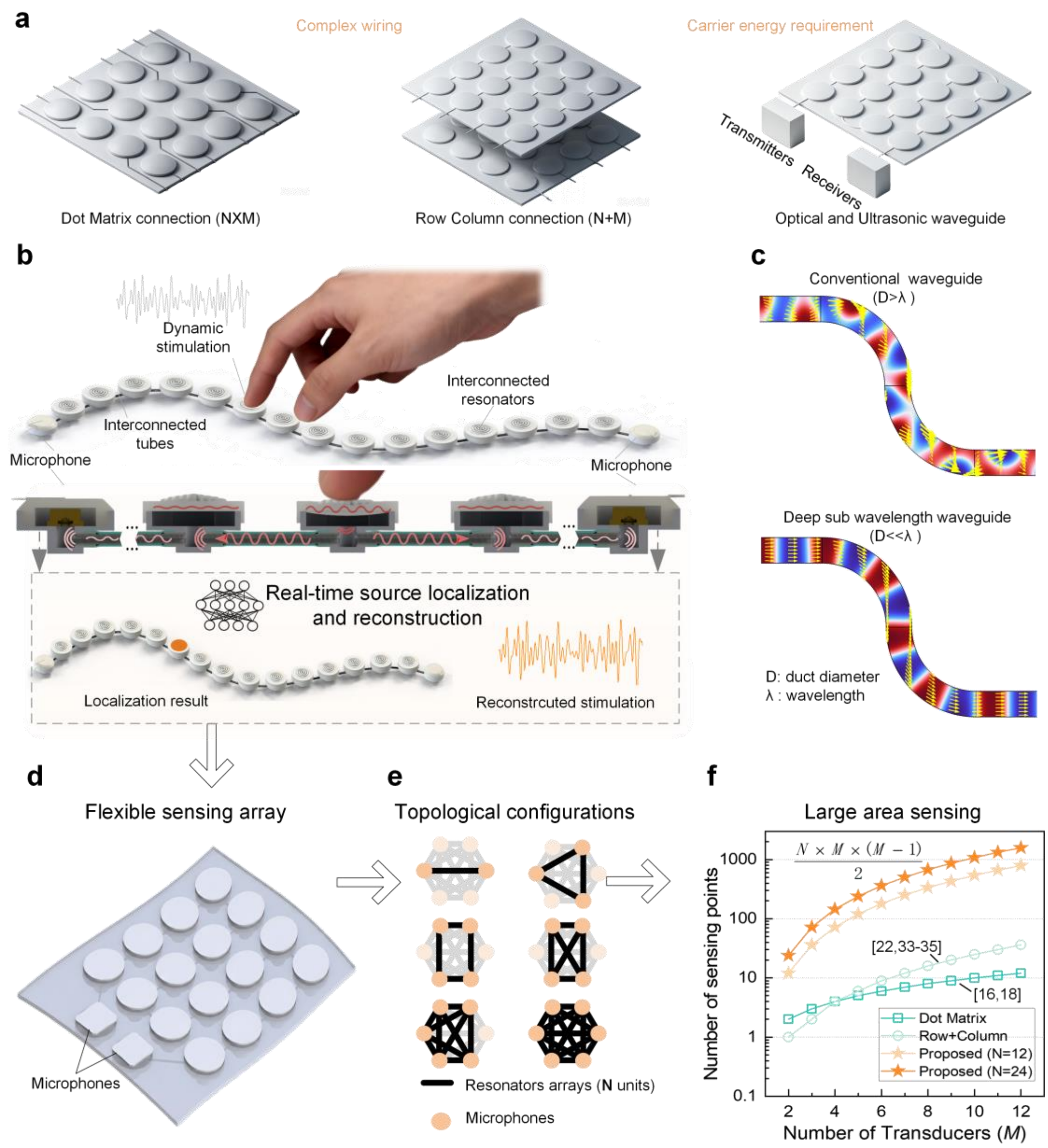

**Fig. 1 | Mechanism of deep-sub-wavelength waveguide dynamic tactile sensing.** a, Schematic of existing large-area tactile arrays. Pixelated arrays suffer from complex wiring; multiplexing only partially alleviates this burden. Distributed waveguide sensing requires active energy injection. b, Sensing principle of the proposed system. Helmholtz resonators capped with flexible membranes are serially interconnected via compliant microtubes to form an acoustic waveguide network. External stimuli are detected by capturing perturbations in the enclosed acoustic field, enabling real-time sound-source localization and stimulus reconstruction. c, Deep-sub-wavelength waveguides under bending. Consistent waveguide behavior is preserved across varying curvatures. d, Schematic of the proposed flexible 2D array. A 1D waveguide array can be coiled into a 2D configuration. e, Configurable topologies. Interconnecting microphone pairs with resonator networks enables diverse configurations by varying both the number and interconnection pattern of microphones. f, Wiring complexity comparison. The number of sensing points scales with both microphone and resonator counts, offering significantly reduced wiring complexity compared with pixelated arrays.

To provide practical guidance for sensor design, we systematically examined the influence of key structural parameters on feature distinguishability using finite-element analysis (see Section 1 of Supplementary Discussion). Key findings indicate that, stimuli applied at each resonator, the waveguide network exhibited clearly distinguishable inter-channel features, particularly in the phase domain. Besides, it is also found that larger resonator cavity volumes and longer interconnecting tubes promote more distinguishable low-frequency characteristics, whereas smaller tube diameters and lower equivalent stiffness of the top plate can further enhance feature distinctiveness. It should be noted, however, that feature distinguishability is not the sole design criterion; sensitivity and structural compactness must also be balanced. For instance, although increasing waveguide impedance improves feature differentiation in the low-frequency range, it may simultaneously introduce greater manufacturing challenges and higher transmission loss, thereby reducing overall sensing sensitivity. These insights offer a quantitative basis for trading off spatial resolution, mechanical compliance, and sensitivity when scaling the architecture to diverse application domains.

## Real-time low-frequency waveguide sound localization

Accurate real-time localization of low-frequency acoustic sources within an enclosed waveguide poses substantial signal-processing challenges, particularly when scaling to high sensing-point densities. To address this, we developed an integrated signal-processing pipeline that combines the fast Continuous Wavelet Transform (fCWT) [19] with a FasterNet backbone [20] (Fig. 2a,b). With large volume simulation dataset (Method and Supplementary Fig. 1), Systematic ablation study are conducted to optimize all hyperparamete (Section 2.1 of Supplementary Discussion).

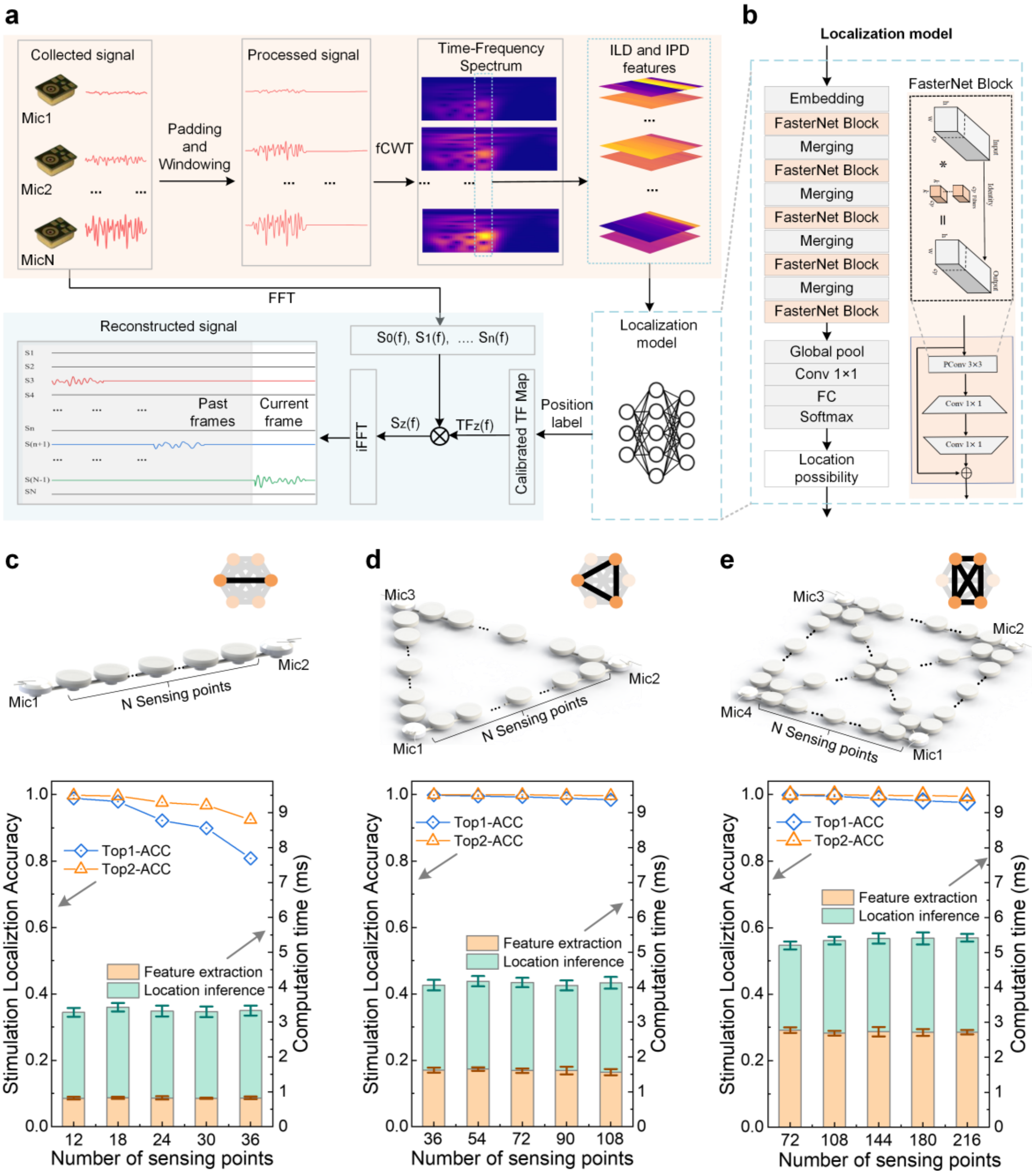

**Fig. 2 | Real-time low-frequency waveguide sound localization.** a, Pipeline for stimulation localization and reconstruction. fCWT is used for multi-channel signal time-frequency analysis. Inter-channel level difference (ILD) and phase difference (IPD) features are extracted for stimulus localization. Signal reconstruction is subsequently performed using the identified location and its calibrated transfer function. b, Stimulation localization model. The input embedding dimension is set to 12. The network comprises five serial FasterNet blocks with depths of [1, 1, 2, 2, 2]. The output is a probability distribution representing localization likelihood. c-e, Performance of the proposed signal processing scheme for sensors with two, three and four microphones, respectively. Adding more transducers primarily increases feature extraction time with minimal impact on single-frame prediction time (2.5~3 ms). Within the same configuration, increasing sensing points reduces accuracy but has little influence on computation efficiency. Configurations with more transducers achieve higher accuracy and support a greater number of sensing points.

Overall, the finalized signal processing pipeline is structured as follows. All microphone signals are sampled at 1000 Hz and segmented into overlapping 250-sample frames with a stride of 16. To mitigate edge effects in time–frequency analysis, each frame is padded by replicating its final sample 250 times. The padded frames are then processed by the fCWT (fast Continuous Wavelet Transform) module to obtain efficient time–frequency decompositions, where the 1–100 Hz target band is uniformly divided into 64 frequency bins. Following feature smoothing, the most recent 32 ms of ILD and IPD features are extracted, stacked across channels, and passed to the neural network for stimulus localization inference. The network comprises an input embedding block of dimension 12, followed by five FasterNet blocks with depths [1, 1, 2, 2, 2], each accompanied by an emerging block, and outputs a probability distribution over all candidate locations.

Using the optimized architecture, and based on finite element simulation, we first characterized the two-microphone configuration to investigate the achievable sensing-point density. As shown in Fig. 2c, for a two microphones system, the Top-1 accuracy remains above 90% when the number of sensing points is below 30. At 36 sensing points, the Top-1 accuracy decreases below 90%, whereas the Top-2 accuracy stays high at 92.5%, indicating reliable localization even at increased densities.

We subsequently extended the system to three- and four-microphone configurations while adopting the structural parameters established in Section 2.2 of Supplementary Discussion. As illustrated in Fig. 2d,e, the additional microphones provide richer discriminative features, enabling both configurations to achieve accuracies exceeding 97%. Specifically, with 36 sensing points per line, the three-microphone system (108 points total) reaches approximately 98.5% accuracy, and the four-microphone system (216 points total) achieves approximately 97%. This modest decrease can be attributed to the doubling of classification categories (from 108 to 216), which increases the complexity of the discrimination task despite the richer inter-channel features provided by the additional microphone. Notably, increasing the microphone count primarily adds to feature extraction time, while model inference efficiency remains largely unaffected. For the three configurations, total time to complete localization of single frame signal respectively below 3.5, 4.5 and 4.5 ms, fully satisfying real-time operational demands.

## Sensor fabrication and validation

This sensing scheme leverages the bending robustness of deep-subwavelength waveguides to decouple sensing performance from array deformation. To validate this, we characterized waveguide tubes comprising an inner micro-spring tube sheathed with an outer PE heat-shrink layer. In the experimental setup (Fig. 3a), a 120-mm-long microtube with an inner diameter of 0.5 mm connected microphone 1 (Mic1) to the excited resonator, while a calibration microphone mounted at the resonator base recorded the acoustic pressure generated by external stimuli. The tube was wound around cylinders of 11.6 mm diameter at bend angles of 0°, 90°, 180°, 270° and 360°. Under each condition, frequency-swept excitation was applied to the resonator, and time–frequency analysis of the two microphone signals yielded the waveguide transfer characteristics. The amplitude–frequency and phase–frequency responses under all deformation states are presented in Fig. 3b,c, demonstrating consistent low-frequency acoustic behavior regardless of bending. This insensitivity to deformation ensures the flexible applicability of the proposed sensing approach.

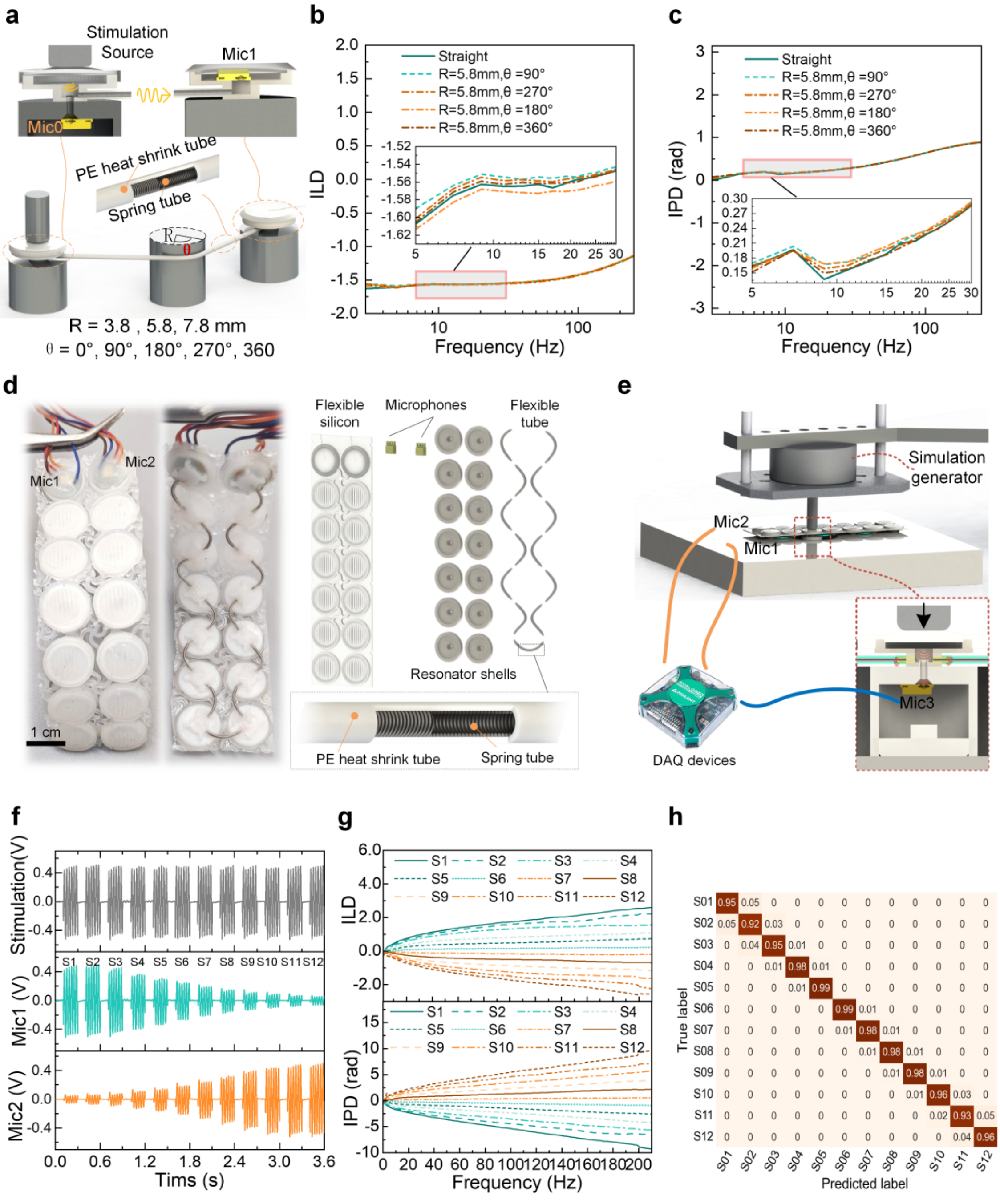

**Fig. 3 | Sensor fabrication and validation**. a, Setup for flexible waveguide robustness characterization. Flexible microtubes are bent to various angles and characterized under swept-frequency excitation. b, c, Frequency-domain amplitude and phase responses of the curved interconnected tube. Highly consistent frequency responses validate the waveguide robustness of the fabricated tubes. d, Fabricated dual-microphone tactile sensor and its detailed composition. e, Schematic of the sensor calibration test. Calibration is performed by applying swept-frequency excitation at individual sensing points while synchronously recording the microphone responses. f, Sensor time-domain response under 31-Hz stimulation. g, Ideal inter-channel level difference (ILD) and inter-channel phase difference (IPD) characteristics obtained from dual-microphone sensor calibration, showing distinct signatures for

different sensing points. h, Confusion matrix of the stimulus-localization model, achieving an average localization accuracy of 96.3%. Most misclassifications occur between adjacent sensing points.

Following the fabrication procedure detailed in Method and Supplementary Fig. 2, we fabricated a sensor with 12 sensing points arranged in a 2 × 6 array (Fig. 3d). The resonator bodies were fabricated by 3D printing, and the flexible top membranes were cast from silicone. The waveguide tubing incorporates a tightly wound internal spring for structural support, encased by a PE heat-shrink sheath. The sensor was calibrated following the setup outlined in Fig. 3e (detail see Method and Supplementary Fig. 3). The responses of different sensing points to an identical 31 Hz stimulus are shown in Fig. 3f. Using swept-frequency stimulation, we obtained the calibrated transfer functions from each sensing point to the two microphones (Supplementary Fig. 4), and then derive the corresponding ILD and IPD characteristics in Fig. 3g. Each sensing point exhibits clearly distinguishable characteristic curves, confirming the spatial discriminability required for accurate localization.

To train the localization model for the physical sensor, we constructed a dataset comprising both numerically simulated and experimentally measured signals acquired under diverse stimulus conditions (see Method). The confusion matrix of the trained model is presented in Fig. 3h, showing an accuracy of 96.3% after 30 training epochs. Real-time demonstrations under various stimulation scenarios—including low-frequency stimulus, contact by different objects, and operation under varying deformation states—further confirm the robustness and adaptability of the system (Supplementary Video 1).

## Intrinsic capabilities and advantages

The scalability of the proposed sensing method also enables high-spatial-resolution tactile perception. With a spatial resolution of 4 mm, we fabricated two fingertip-scale dual-microphone sensors by arranging resonators into 3 × 4 (Fig. 4a and Supplementary Fig. 5) and 6 × 4 arrays (Supplementary Fig. 6). As illustrated in Fig. 4b, the sensor demonstrates ultra-high sensitivity, capable of detecting the gentle touch of a single human hair (Supplementary Videos 2).

Leveraging the topological configurability of the sensing approach, we further fabricated a three-microphone tactile glove (Fig. 4c and Supplementary Fig. 7) and a four-microphone large-area sensing array (Fig. 4d and Supplementary Fig. 8). Both arrays exhibit excellent structural flexibility. After calibrations (Supplementary Fig. 9,10), corresponding stimulus-localization models are trained for these sensors (Supplementary Fig. 11,12). As shown in Fig. 4e, the system consistently achieves high localization accuracy across diverse array topologies incorporating large numbers of sensing points. Furthermore, increasing the number of microphones expands the total sensing-point count and enhances localization accuracy, with certain configurations exceeding 98% accuracy.

Given the use of microphones as transducers, we also investigated the influence of environmental noise on sensing performance. The experimental setup for noise-signal collection is shown in Fig. 4f: a speaker emitting industrial noise was positioned adjacent to the smart glove, and the glove's response was recorded under varying ambient noise levels. The results indicate that the glove's response to environmental noise is minimal. The signal-to-noise ratio (SNR) at different noise levels is shown in Fig. 4g; even at 90 dB ambient noise—where the microphone output is approximately 0.01 V—the sensor maintains a signal quality corresponding to an SNR of 10 dB. To further assess robustness, we added ambient noise to the original dataset to construct a noise-contaminated dataset (Supplementary Fig. 13), which was then used to evaluate the trained model. As shown in Fig. 4h, no significant degradation in localization accuracy is observed even at noise levels up to 90 dB. A demonstration of system performance under different noise conditions is provided in Supplementary Video 3.

Beyond localization, the practical utility of the sensor is further enhanced by stimulus signal reconstruction. Details of the reconstruction process are provided in Fig. 4i. Examples of reconstructed time-domain signals are shown in Fig. 4j and Supplementary Fig. 14,15. Evaluating reconstruction quality using the signal-to-noise ratio (SNR), Fig. 4k shows that increasing the short-time Fourier transform (STFT) window length improves reconstruction fidelity. Using a 32-ms window—which is sufficient for many practical applications—yields a minimum average SNR of 7.73 dB. Moreover, sensing points with higher intrinsic sensitivity exhibit superior reconstruction performance, owing to the higher SNR of the raw microphone signals used in the reconstruction process.

By combining localization and reconstruction, the sensing system operates at frame rates exceeding 60 fps. Even when running a real-time interactive demonstration interface—which substantially reduces computational efficiency—the system maintains a response rate above 30 fps (Supplementary Video 1). In summary, the proposed platform delivers robust and reliable performance in both stimulus localization and reconstruction, while fully meeting the stringent requirements for real-time operation.

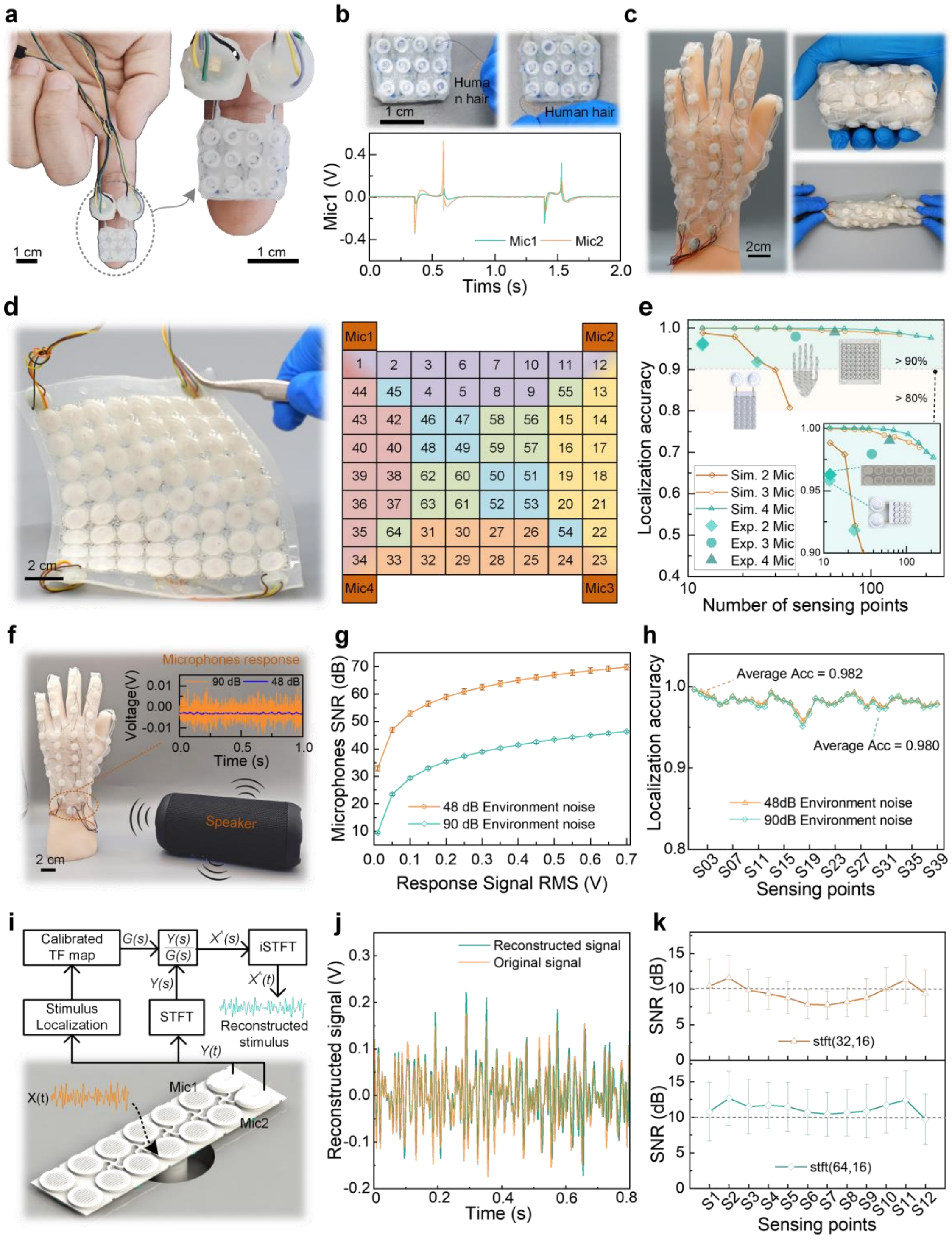

**Fig. 4 | Intrinsic capabilities and advantages of proposed sensing method**. a, Fabricated dual-microphone-based tactile sensor with high spatial resolution; a 3x4 array achieves a spatial resolution of 4 mm. b, High sensitivity: the sensor detects stimulation from a single human hair. c, Fabricated smart glove, which flexible enough to withstand severe twisting and rolling deformations. d, Fabricated large-area sensor incorporating four microphone, with numbering of its 64 sensing points. e, High accuracy:

high accuracy is sustained across all tested configurations. f, Experimental setup for environment noise testing, in which a speaker emitting noise was placed adjacent to the glove to evaluate noise interference. g, Theoretical signal-to-noise ratio across varying environment noise levels. h, Localization accuracy under varying noise conditions, showing negligible influence of noise. i, Schematic for stimulus reconstruction. j, Time-domain reconstruction of a dynamic excitation signal, with STFT window as (32,16). k, Signal-to-noise ratio (SNR) of reconstructed stimulation, with a minimum average SNR of 7.73 dB.

## Application demonstration

This sensing system finds diverse applications in scenarios requiring dynamic tactile detection. As a proof-of-concept demonstration, a fingertip-scale 3 × 4 sensor was integrated into a Whack-a-Mole interactive interface (Fig. 5a and Supplementary Video 2). Moreover, the sensor can be configured as wearable electronic skin, serving as a synthetic counterpart to biological dynamic mechanoreceptors for detecting subtle external contacts. As illustrated in Fig. 5b and Supplementary Video 3, the system exhibits ultra-high sensitivity by resolving gentle contacts with soft paper towels and feathers. Furthermore, the tactile glove can be applied to pulse-wave sensing. Fig. 5c shows the raw signals captured by the three microphones when measuring arterial pulse waves at the ring fingertip, with more results in Supplementary Fig. 16 and Supplementary Video 3. Notably, reconstruction quality varies across microphones due to differences in signal-to-noise ratio (SNR); however, for each sensing point, at least one channel consistently exhibits sufficiently high SNR to enable high-fidelity reconstruction. This capability permits detection of not only the primary pulse wave but also secondary features such as the tidal and dicrotic waves, supporting advanced pulse-wave analysis and potential diagnostic applications.

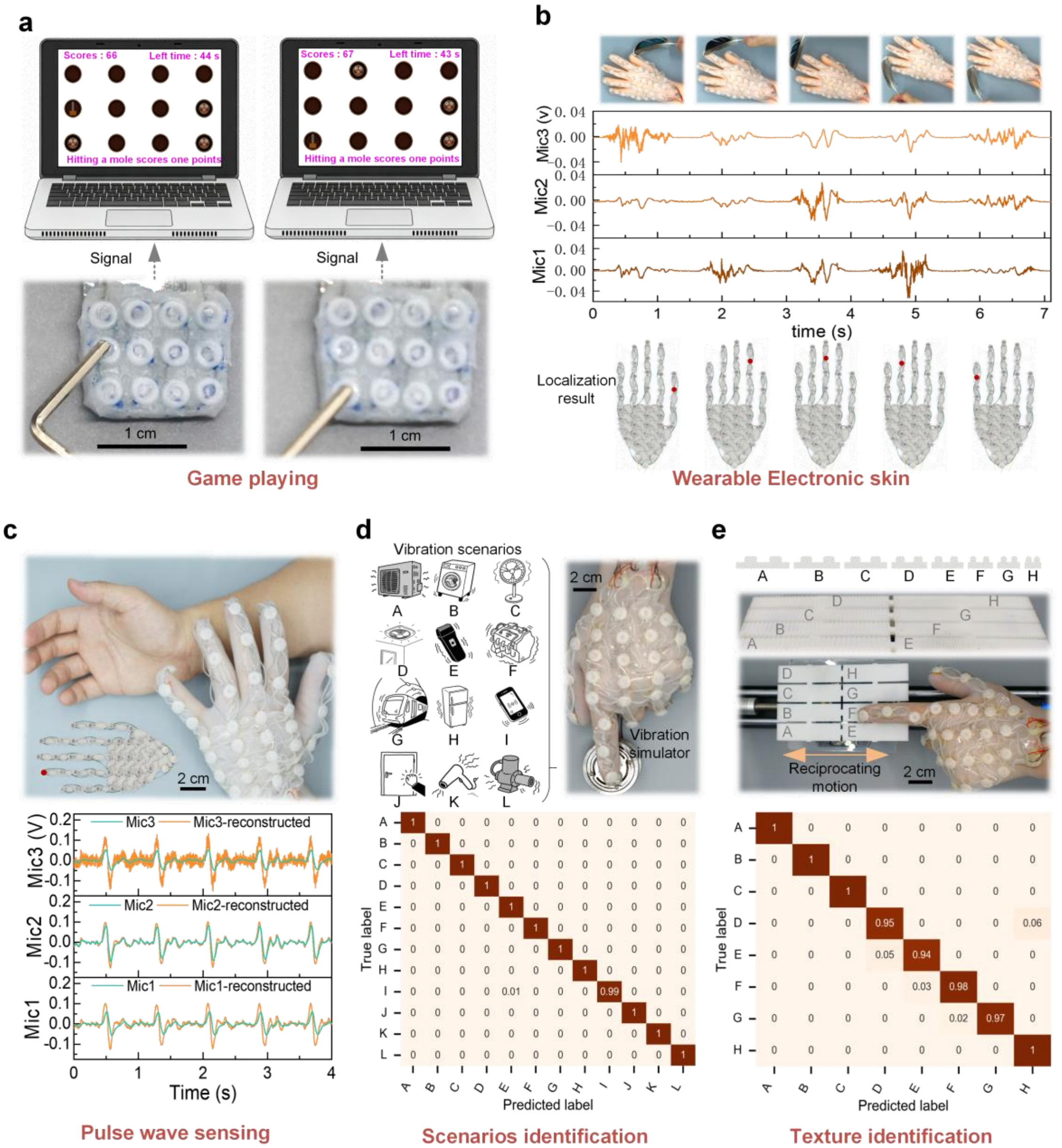

Fig. 5 | **Application demonstration (Part-1)**. a, Game playing. Real-time Whac-A-Mole gameplay using the high-resolution dual-microphone-based tactile sensors, demonstrating rapid localization of low-frequency stimuli with 4-mm spatial resolution. b, Wearable electronic skin. The dynamic tactile glove applied for the detection of subtle feathers contact. c, Pulse wave sensing. Tactile glove applied for wrist pulse wave sensing, demonstrates capability for detection and reconstruction of ultra-low frequency stimulus. d, Vibration event identification. Achieving an accuracy over 99% for identification of 12 scenarios in daily life. e, Textures identification Achieving an accuracy over an 97.9% for identification of 8 different textures.

Using the workflow outlined in Supplementary Fig. 17**a**, the tactile glove can also perform stimulus-modality identification. First, it was applied to event recognition, specifically classifying 12 common low-frequency vibration events from daily life (Fig. 5d; dataset details are available in Supplementary Fig. 17e–p). A recognition accuracy of 99.81% was achieved, as shown in the confusion matrix in Fig. 5d, with further demonstrations in Supplementary Fig. 18 and Supplementary Video 3. Subsequently, the glove was applied to texture recognition. Eight 3D-printed textures with distinct geometric dimensions were prepared (Fig. 5e**;** details in Supplementary Fig. 17**b–c**). As the finger slid across each texture at 1.2 cm/s, characteristic template profiles were obtained for all surfaces (Supplementary Fig. 17**d**). Using the same recognition scheme, the system achieved a texture-identification accuracy of 97.92%, with the corresponding confusion matrix shown in Fig. 5e and functional demonstrations in Supplementary Fig. 19 and Supplementary Video 3. Importantly, incorporating these downstream tasks does not compromise real-time performance: averaged over 5,000 cycles, inter-channel feature extraction requires 1.549 ms, stimulus localization 2.661 ms, and modality feature extraction and classification only 1.151 ms. The combination of high accuracy and low computational overhead confirms the glove's suitability for complex real-world applications.

The sensing system also extends to environmental monitoring and spatial perception. As shown in Fig. 6a**,** the large-area sensor reliably detects impacts from a 5-mg particle falling from heights ranging from 1 cm to 4 cm. Supplementary Video 4 demonstrates that the sensor simultaneously achieves large-area coverage, high accuracy, and high sensitivity in capturing such lightweight objects. The large-area sensor was further applied to stimulus-trajectory tracking and subsequent trajectory-pattern recognition. By performing rule-based analysis on the localization records (Supplementary Fig. 20**a**), we reconstructed the historical stimulus trajectories. Fig. 6b,c and Supplementary Fig. 20b,c,d presents examples of trajectories captured under various conditions, including a rolling ball, a crawling insect, and air blowing.

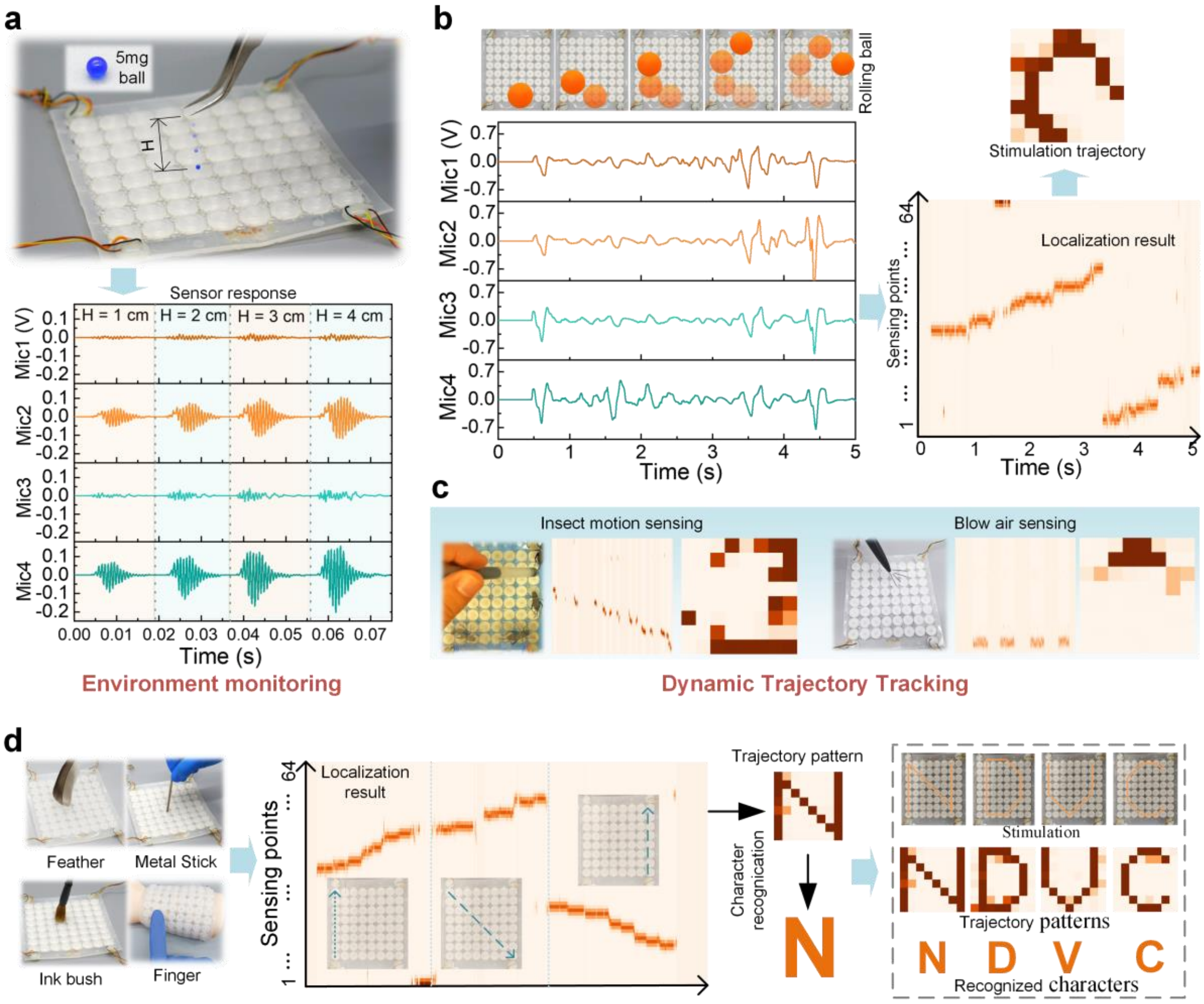

Fig. 6 | **Application demonstration (Part-2)**. a, Environment monitoring: the large area sensor applied as a intelligent carpet for detection of a 5 mg object dropped from a height of 1-4 cm. **b,** Dynamic tracjectory tracking: The large area sensor used for tracking of rolling ball**. c**, Dynamic trajectory tracking: the large area sensor used for incest tracking and behavior studied, as well as detection of airflow detection. **d**, Flexible sliding-touch input interface. Based on stimulus trajectory tracking, the sensor can be further applied as an interactive panel for inputting texts or commands.

By integrating captured dynamic-contact trajectories with downstream models, the sensing system enables more complex stimulus-behavior analysis. To demonstrate this, a neural network comprising two FasterNet blocks (with depths of 1 and 2) was trained to recognize 26 capital letters. The training dataset was generated by writing characters on the sensor surface and recording the corresponding trajectory patterns. For each character, 15 min of time-domain data were collected—yielding approximately 60 samples—and augmented with salt-and-pepper noise. After 30 training epochs, the model achieved a character-identification accuracy exceeding 99%, with the confusion matrix shown in Supplementary Fig. 21. Fig. 6d provides an example of the real-time letter-recognition process, illustrating the stimulation path, the reconstructed trajectory, and the predicted character. As demonstrated in Supplementary Video 4**,** sliding motions produced by a feather, a metal stick, a soft ink brush, and a human finger are precisely captured and identified.

## Discussion

This study establishes a sensing paradigm that integrates acoustic metamaterials with tactile perception, enabling the detection of large-area dynamic stimuli with minimal hardware complexity. The core advances are twofold. First, by exploiting deep-subwavelength acoustic waveguides, we realize an acoustic metamaterial structure whose transmission characteristics are inherently immune to macroscopic deformation. This decoupling arises because the operating wavelength far exceeds the waveguide cross-section, forcing the wave to propagate in a quasi-plane-wave mode whose dispersion relation is insensitive to curvature. To the best of our knowledge, such intrinsic mechanical robustness is unprecedented in waveguide-based tactile systems. Second, through the synergistic coupling of time–frequency analysis and lightweight neural-network classification, we achieve real-time millimeter-scale localization of dynamic excitations over the 1–100 Hz band—a regime where conventional beamforming or time-difference-of-arrival (TDOA) methods fail due to diffraction-limited spatial resolution at low frequencies. This performance stems not from algorithmic innovation alone but from the tight integration of waveguide physics and signal processing: the metamaterial front end sculpts the acoustic field into a representation that the back end can decode with exceptional efficiency.

The ultra-high sensitivity of the scheme originates from the strong energy confinement within enclosed, deep-subwavelength channels. Proof-of-concept experiments demonstrate that the system produces a clear, distinct response to the impact of 5 mg particles dropped from a height of merely 1 cm—corresponding to a mechanical impact energy of approximately 0.5 μJ—as well as to the delicate contact of a single human hair or a feather. These observations confirm the system's ability to perceive mechanical stimuli at extremely low energy thresholds, highlighting its promise as an ultrasensitive tactile sensing platform. Beyond mere detection, the waveguide architecture faithfully preserves the temporal structure of propagating waves, enabling real-time reconstruction of dynamic excitation waveforms with a signal-to-noise ratio (SNR) exceeding 10 dB at every sensing point. This capability fundamentally distinguishes the method from threshold-based binary sensors, providing the continuous tactile information essential for dexterous manipulation and texture recognition. Critically, the same deep-subwavelength confinement mechanism that confers this sensitivity also underpins the system's mechanical compliance.

The mechanical resilience of the system is a direct consequence of the same deep-subwavelength physics. Because the acoustic mode is tightly confined and its effective wavelength far exceeds the radii of curvature encountered during coiling, wrapping, or severe bending, macroscopic deformation perturbs neither the local dispersion relation nor the accumulated modal phase. Experiments show that a waveguide with a 0.5 mm inner diameter, bent to a radius of 5.8 mm, maintains stable waveguiding with negligible transmission degradation for acoustic waves below 100 Hz. Consequently, both the hardware and software architectures support topological adaptability: the array can be coiled, wrapped around curved surfaces, or arbitrarily reshaped without recalibration. We demonstrate this universality by fabricating a three-microphone tactile glove and a four-microphone large-area array, both of which maintain stable performance under diverse hand gestures and deformation conditions. This topological adaptability is realized by simply adjusting the number of microphones and sensing units, allowing the same modular design to seamlessly scale from a fingertip-sized patch to large-area skin.

With regard to spatial resolution and scalability, fabricated prototypes achieve a localization accuracy as fine as 4 mm, a figure that can be further refined with advanced fabrication. As the number of microphones increases, the total number of classifiable sensing points scales as $N \times M(M-1)/2$, where N is the number of sensing points per line and M is the number of microphones, enabling high-density coverage without a proportional increase in hardware complexity. Although the expanded classification space elevates discrimination difficulty, the richer inter-channel features introduced by additional microphones sustain high performance; the system achieves ~97% top-1 accuracy for 216 points with four microphones, compared to ~98.5% for 108 points with three. Notably, these sensors readily attain nearly 100% top-2 accuracy, indicating that the topological redundancy afforded by extra microphones enhances localization robustness—a key advantage when approximate position information suffices. Ambient noise tests further reveal that the enclosed waveguide architecture not only ensures high sensitivity but also intrinsically isolates the sensing space from external acoustic interference, rendering performance largely impervious to environmental noise. Moreover, the fabrication process requires no specialized equipment; laboratory-scale prototypes cost from a few dollars to approximately 15 USD, a figure poised to drop dramatically upon industrial-scale production.

When compared with existing technologies, the trade-offs of this paradigm become clear. Conventional discrete tactile arrays require dense, element-specific wiring that scales linearly with the number of sensing points, fundamentally limiting scalability and yield in large-area implementations. Our approach collapses this wiring complexity by multiplexing numerous resonators into a few microphones via the waveguide network, trading deterministic pixel addressing for probabilistic acoustic-source classification. Consequently, the framework prioritizes coverage, mechanical compliance, and hardware efficiency over error-free point-to-point mapping. Unlike active-interrogation waveguide systems that rely on externally injected energy (e.g., optical or ultrasonic modalities), our strategy is entirely passive-listening, directly detecting the acoustic signals generated by the stimulus itself. This eliminates the need for onboard lasers or ultrasonic transceivers, but also implies that detection sensitivity is bounded by the acoustic energy produced by the excitation rather than by the signal-to-noise ratio of an active interrogation scheme. Crucially, existing waveguide sensors suffer from mode coupling and phase distortion under bending because their transmission characteristics are geometry-dependent. The deep-subwavelength framework fundamentally circumvents this by confining all waveguide behavior to a quasi-one-dimensional plane-wave mode, rendering the propagation characteristics intrinsically impervious to macroscopic curvature.

The versatility of this paradigm is validated across diverse operational scenarios, including interactive interfaces, physiological monitoring, environmental sensing, and trajectory analysis. These proof-of-concept demonstrations collectively establish a platform that simultaneously supports stimulus localization, modality identification, and dynamic waveform reconstruction—capabilities rarely unified within a single passive tactile architecture. While the present study focuses on the 1–100 Hz band, the underlying deep-subwavelength physics is not intrinsically limited to this regime; in principle, the same sensing scheme can be extended to any frequency range and waveguide scale where the deep-subwavelength condition is satisfied. Beyond these validated functions, the mechanical compliance and scalability of the architecture suggest immediate potential in wearable health monitoring (e.g., respiration or muscle tremor detection) and robotic manipulation (e.g., incipient slip detection in soft grippers), where large-area conformable tactile sensing is critical.

At this proof-of-concept stage, several limitations remain. First, the current framework assumes a single excitation source at any given time. Overlapping time–frequency signatures from simultaneously active sources cannot be unambiguously resolved with the present spectrogram-based representation. Extending this capability will likely require algorithmic innovations, such as sparse source separation or learned dictionaries capable of deconvolving superimposed acoustic events. Second, the operating bandwidth has fundamental boundaries. At the low-frequency limit, sensitivity is ultimately constrained by microphone self-noise and the thermo-mechanical noise floor of the elastomeric waveguide, which together define the minimum detectable acoustic displacement. We have validated performance within 100 Hz but have not systematically explored the full frequency spectrum permissible under the deep-subwavelength condition. Third, the long-term material stability under cyclic loading—including repeated coiling, stretching, and exposure to humidity or temperature variations—requires systematic characterization. Finally, although the signal-processing pipeline currently operates on a host computer for algorithmic flexibility, real-time untethered operation necessitates porting the time–frequency analysis and lightweight classification models to dedicated microcontroller units. Future work will address these limitations, aiming to extend multi-source capability and bandwidth through algorithmic upgrades while migrating the complete pipeline to embedded hardware.

In summary, this study introduces a distinctive technical paradigm that leverages the unique physics of deep-subwavelength acoustic confinement to simultaneously achieve unparalleled sensitivity, mechanical flexibility, and large-area scalability. By fundamentally decoupling sensing performance from mechanical deformation, our framework paves the way for a new class of tactile skins that can conform to arbitrary geometries without sacrificing perceptual fidelity. This tight integration of physical intelligence and computational decoding marks a promising direction for the next generation of soft, perceptive robotic systems and bio-integrated electronics.

# Method

## Sensor fabrication

The SPH1878LR5H-C MEMS microphone (Knowles Corporation) served as the acoustic transducer. The overall fabrication sequence is illustrated in Supplementary Fig. 2. Each sensing resonator consists of a rigid shell and a pre-fabricated silicone rubber membrane: the shell was 3D-printed from a photosensitive resin, while the membrane was cast in 3D-printed molds. The resonators and micro-tubes were then assembled into an array. Airtight connections between the micro-tubes and the rigid shells were achieved using a liquid adhesive, and the shells were bonded to the top membranes with silicone glue. Finally, the MEMS microphones were embedded according to the desired configuration and encapsulated with silicone adhesive. This process is applicable to all sensor variants based on the reported sensing principle. The material specifications and critical dimensions of the prototypes used in this study are given below.

For the two-microphone $2 \times 6$ array sensor (Fig. 3d), the elastic top membrane was fabricated from liquid silicone rubber with a Shore A hardness of 30. Each interconnected micro-tube segment measured 12.5 mm in length and was constructed by inserting a fully sealed spring tube (wire diameter 0.08 mm, outer diameter 0.6 mm) into a polyethylene (PE) heat-shrink tube that had a pre-shrink inner diameter of 0.9 mm and a wall thickness of 0.15 mm.

For the high-spatial-resolution sensors, two configurations were fabricated. In the 3 × 4 array (Supplementary Fig. 5), the elastic top cover was formed from a 5:1 blend of two liquid silicone rubbers with Shore A hardness values of 0 and 5. In the 4 × 6 array (Supplementary Fig. 6), the top elastic plate was fabricated from liquid silicone rubber with a Shore A hardness of 5. In both sensors, each flexible micro-tube segment was 4 mm long and comprised a fully sealed spring tube (wire diameter 0.05 mm, outer diameter 0.28 mm) sheathed in a polyurethane soft catheter (outer diameter 0.55 mm, inner diameter 0.28 mm). The two microphones were connected by waveguide tubes, each 12 mm in length, that were made from a fully sealed spring tube (wire diameter 0.08 mm, outer diameter 0.6 mm) enclosed in a PE heat-shrink tube with a pre-shrink inner diameter of 0.9 mm and a wall thickness of 0.1 mm.

The tactile glove sensor had detailed dimensions as shown in Supplementary Fig. 7. The elastic top membrane was fabricated from liquid silicone rubber with a Shore A hardness of 30. The interconnected micro-tubes were made from PE heat-shrink tubing that had a pre-shrink inner diameter of 0.9 mm and a wall thickness of 0.15 mm, while the embedded spring tubes possessed an outer diameter of 0.8 mm and a wire diameter of 0.08 mm.

The large-area sensor had detailed dimensions provided in Supplementary Fig. 8. The elastic top membrane was fabricated from liquid silicone rubber with a Shore A hardness of 30. All interconnected micro-tubes were constructed from PE heat-shrink tubing with an inner diameter of 0.9 mm and a wall thickness of 0.15 mm. The spring tubes highlighted in yellow and green in the back-side view of Supplementary Fig. 8a had outer diameters of 0.7 mm and 0.8 mm, lengths of 12 mm and 14 mm, respectively, and both shared a wire diameter of 0.08 mm.

## Sensor calibration

For sensor calibration, a platform shown in Supplementary Fig. 3, where a voice-coil motor controlled via Bluetooth from a smartphone served as the excitation source. During calibration, the smartphone played a predefined swept-frequency signal (1–125 Hz), prompting the motor to generate corresponding vibrational excitation. A calibration microphone attached to the base of the excited resonator captured the original acoustic signal. By collecting responses from all three microphones, the numerical acoustic transfer function from each sensing point to the two sensing microphones was derived.

## Signal processing

The features used for acoustic localization are the inter-channel level difference (ILD) and inter-channel phase difference (IPD), which defined as follows:

$$ILD(f,n) = 20log_{10}\left|\frac{X_1(f,n)}{X_2(f,n)}\right| \tag{1a}$$

$$IPD(f,n) = \angle\frac{X_1(f,n)}{X_2(f,n)} \tag{1b}$$

where $X_1(f,n)$ and $X_2(f,n)$ denote the time-frequency representations of the signals received by two microphones.

The target frequency range (1–100 Hz) is linearly divided into 64 frequency bins. From the resulting time–frequency spectra, inter-channel level difference (ILD) and inter-channel phase difference (IPD) features are extracted from the most recent 47 ms of unpadded signal, generating ILD and IPD matrices of size 64 × 47. These matrices are subsequently smoothed using an average filter with a kernel size of (1, 16) and stride 1, producing aggregated representations corresponding to the latest 32 ms. The ILD and IPD matrices are then stacked to form the final localization feature set. For systems with two, three, and four microphones, the resulting feature dimensions are 2 × 64 × 32, 6 × 64 × 32, and 12 × 64 × 32, respectively.

The model was trained using AdamW optimizer and cosine learning rate schedule, with initial rate as 0.001. The dataset was split into training and testing subsets in a 4:1 ratio. To mitigate model overconfidence, Gaussian soft labels were applied by convolving each target label vector with a Gaussian kernel ($\sigma = 1$), transforming the original one-hot distribution into a smooth, continuous-valued distribution that allocates probability mass to neighbouring classes. Training was conducted using the SoftTargetCrossEntropy loss function provided in the timm library.

## Building of model training datasets

All sensor stimulus localization models based on finite element simulation were constructed with simulation. For physically fabricated sensors, in addition to the simulated data generated, real test data were also incorporated. Detail about both simulates data and real test data are given below.

(1) Generation simulation data for model training

Based on the finite element analysis and/or sensor calibration, the obtained sensor amplitude-frequency and phase-frequency characteristics can be expressed as:

$$A_{mi}(f) = \frac{|S_{mi}(f)|}{|S_i(f)|} \tag{2}$$

$$P_{mi}(f) = \angle S_{mi}(f) - \angle S_i(f) \tag{3}$$

where $f$ is the corresponding frequency bin, $S_i(f)$ is the sound signal generated at the stimulated sensing points $i$. $S_{mi}(f)$ is the corresponding response of microphone $m$.

The frequency domain numerical transfer function can be expressed as:

$$TF_{mi}(f) = A_{mi}(f)e^{j*P_{mi}(f)} \tag{4}$$

where the dimension of $f$ is $1 \times n$, and a transfer vector $TF_{mi}(f)$ composed by discrete complex values, was obtained for each sensing point-microphone pair. By further simulating a series of virtual excitation signals $S_i(f)$, these complex vectors enable efficient generation of large-scale simulated datasets according to the following equation:

$$S_{mi}(f) = TF_{mi}(f) * S_i(f) \tag{5}$$

To ensure exact frequency alignment between the transfer vectors and the short-time Fourier transform (STFT) result of the virtual stimulation $S_i(f)$, the original numerical transfer function is interpolated along the frequency axis during implementation. With the microphone responses in the frequency domain $S_{mi}(f)$ obtained, inverse STFT was performed to obtain the microphone signals in the time-domain. A diagram about this data generation process is given **Supplementary Fig. 1**.

The time domain virtual excitation signals $S_i(t)$ were formed through the superposition of $N$ sinusoidal components, expressed as:

$$S(t) = \sum_{i=1}^{N} A_i \, sin(2\pi f_i t + \theta_i) \tag{6}$$

where the number $N$ is a random value, the frequency $f_i$ is set to the range of 1–100 Hz, and the amplitude $A_i$ and phase $\theta_i$ are uniformly drawn from [0.02, 0.1] and [-π, π], respectively. To avoid signal saturation caused by an overly high amplitude, the obtained time-domain signal was normalized as follows:

$$S_i(t) = \begin{cases} S_i(t) + n(t) & if \quad Max(|S_i(t)|) < 1 \\ \frac{S_i(t)}{Max(|S_i(t)|)} + n(t) & if \quad Max(|S_i(t)|) > 1 \end{cases} \tag{7}$$

where n(t) is additive white Gaussian noise used to make the signal more like real stimulation. The mean value of the noise is 0, while the standard deviation is set to 0.005.

To maximize coverage of potential frequency combinations within the 1–100 Hz band using limited data, simulated excitation signals were generated by randomly selecting the number of frequency components $N$ in Eq. (6) from seven intervals: [1,4), [4,20), [20,40), [40,60), [60,80), [80,100), and [1,101). This approach models excitations with diverse frequency component counts. In this study, 14,000 s of virtual stimulation data were generated for each sensing point, with 2,000 s allocated to each interval. Assuming one second corresponds to a single stimulation condition, a total of 14,000 simulated feature samples were obtained per sensing point.

(2) Collection of real experimental data for model training

To enhance the dataset for training the stimulation localization model, true sensor responses under four stimulation scenarios were collected. Each stimulation was applied sequentially to every sensing point, and the corresponding data were recorded. Details of the four stimulation scenarios are provided below:

**Part 1 (1–100 Hz Random Excitation)**: 13 minutes signals with random frequencies ( >1Hz and <100Hz ) and amplitudes were pre-generated, and used to drive voice coil motor for sensing point excitation.

**Part 2 (1–10 Hz Random Excitation)**: 2 minutes low-frequency signals were designed to improve model performance in the ultra-low frequency range.

**Part 3 (Finger Friction Excitation)**: 2 minutes recordings of microphone responses to non-harmonic stimuli (brush and friction motions) were collected for each sensing point.

**Part 4 (Tap/Press Excitation)**: Responses to transient finger taps were recorded for two minutes per sensing point

## Author's contributions

G. LONG conducted the research, prepared the materials, and completed all experiments with data analysis; C. LINGHU, C. LIU and K. XU validated the data, discussed and prepared material organization and data analysis; XJ Jing proposed the initial idea, led the research, supervised methodology, provided all funding supports, reviewed and revised the paper.

## Acknowledgement

The work is supported by a NSFC-RGC joint research scheme (N_CityU114/23),

General Research Funds of Hong Kong RGC (11202323; 11204724), a booster fund of City University of Hong Kong (7030015), a Collaborative Research Fund of Hong Kong RGC (C1013-24G), an Innovation and Technology Funds of Hong Kong ITC (ITP/003/24LP, GHP/064/22), and a startup fund from City University of Hong Kong (Ref. 9380140). Changhong LINGHU acknowledges the support by a grant from City University of Hong Kong (Project No. 9382018). We would also like to acknowledge Mr. Zhicheng LIANG from ChangSha, Hunan, China, for his help in providing hardware for model training and evaluation.

## Conflicts of interests

The authors declare no competing interests.

## Additional Information

Correspondence and requests for materials should be addressed to Xingjian JING.